\documentclass[letterpaper, 10 pt, conference]{ieeeconf}  % Comment this line out if you need a4paper
\usepackage{cite}
\usepackage{amsmath,amssymb,amsfonts}
\usepackage{pifont} % For xmark symbol
\newcommand{\xmark}{\ding{55}}
\usepackage{algorithmic}
\usepackage{caption}
\usepackage{subcaption}
\usepackage[ruled,vlined]{algorithm2e}
\usepackage{graphicx}
\usepackage{array}
\usepackage{hhline}
\usepackage{hyperref}
\graphicspath{ {./figures/} }
\usepackage{textcomp}
\usepackage{xcolor}
\usepackage{balance}

\title{Residual Policy Learning for Perceptive Quadruped Control Using Differentiable Simulation \\
{\large Jing Yuan Luo\textsuperscript{1}, Yunlong Song\textsuperscript{2}, Victor Klemm\textsuperscript{1}, Fan Shi\textsuperscript{3}, Davide Scaramuzza\textsuperscript{2}, Marco Hutter\textsuperscript{1}} \\
{\small \textsuperscript{1}ETH Zurich, Switzerland, \textsuperscript{2}University of Zurich, Switzerland, \textsuperscript{3}National University of Singapore}

}

\begin{document}

\maketitle
\thispagestyle{empty}
\pagestyle{empty}

\begin{abstract}

First-order Policy Gradient (FoPG) algorithms such as \textit{Backpropagation through Time} and \textit{Analytical Policy Gradients} leverage local simulation physics to accelerate policy search, significantly improving sample efficiency in robot control compared to standard model-free reinforcement learning. However, FoPG algorithms can exhibit poor learning dynamics in contact-rich tasks like locomotion. Previous approaches address this issue by alleviating contact dynamics via algorithmic or simulation innovations. In contrast, we propose guiding the policy search by learning a residual over a simple baseline policy. For quadruped locomotion, we find that the role of residual policy learning in FoPG-based training (FoPG RPL) is primarily to improve asymptotic rewards, compared to improving sample efficiency for model-free RL. Additionally, we provide insights on applying FoPG's to pixel-based local navigation, training a point-mass robot to convergence within seconds. Finally, we showcase the versatility of FoPG RPL by using it to train locomotion and perceptive navigation end-to-end on a quadruped in minutes.

\end{abstract}

\section{Introduction}

On-Policy Reinforcement Learning (RL) has driven recent advances in quadruped locomotion, enabling the development of highly robust \cite{mikiLearningRobustPerceptive2022} and dynamic \cite{chengExtremeParkourLegged2024} policies. These breakthroughs are primarily powered by on-policy RL algorithms that leverage large-scale data collection to compute zeroth-order policy gradient estimates (ZoPG). ZoPG methods estimate gradients using only reward signals \cite{suhDifferentiableSimulatorsGive2022a}, treating the simulation as a black box and eliminating the need for differentiability. This approach extends RL's applicability to discrete tasks, such as pixel-space control \cite{mnihPlayingAtariDeep2013}, and allows for flexible reward engineering \cite{songReachingLimitAutonomous2023}.

This flexibility, however, comes with the drawback of high gradient variance, which scales poorly with task dimensionality \cite{rudinLearningWalkMinutes2022a}. Although stabler training algorithms \cite{schulmanProximalPolicyOptimization2017, schulmanHighDimensionalContinuousControl2018, haarnojaSoftActorCriticPolicy2018} and hardware acceleration for faster sample collection have mitigated some of these challenges, tasks like vision-based control often still require hybrid approaches. These include training a teacher policy on a simplified setup using RL, which then supervises the final policy \cite{mikiLearningRobustPerceptive2022, heAgileSafeLearning2024, chengExtremeParkourLegged2024}.

In robot control, the underlying dynamics are known, making ZoPG methods overlook valuable local information from dynamics Jacobians. First-Order Policy Gradient (FoPG) methods incorporate these dynamics by directly differentiating returns through the robot's trajectory to each action, including dynamics Jacobians into gradient estimation. Prior works have called this approach \textit{Backpropogation through Time} (BPTT) \cite{wiedemannTrainingEfficientControllers2023a, xuAcceleratedPolicyLearning2022a} and \textit{Analytical Policy Gradients} (APG) \cite{freemanBraxDifferentiablePhysics2021}. This approach can reduce gradient variance \cite{suhDifferentiableSimulatorsGive2022a} but may create a non-smooth loss landscape due to simulation details like contact interactions \cite{antonovaRethinkingOptimizationDifferentiable2023} and rendering artifacts (section \ref{section:diff_thru_pixels}). 

\begin{figure}
    \centering
    \includegraphics[width=0.75\linewidth]{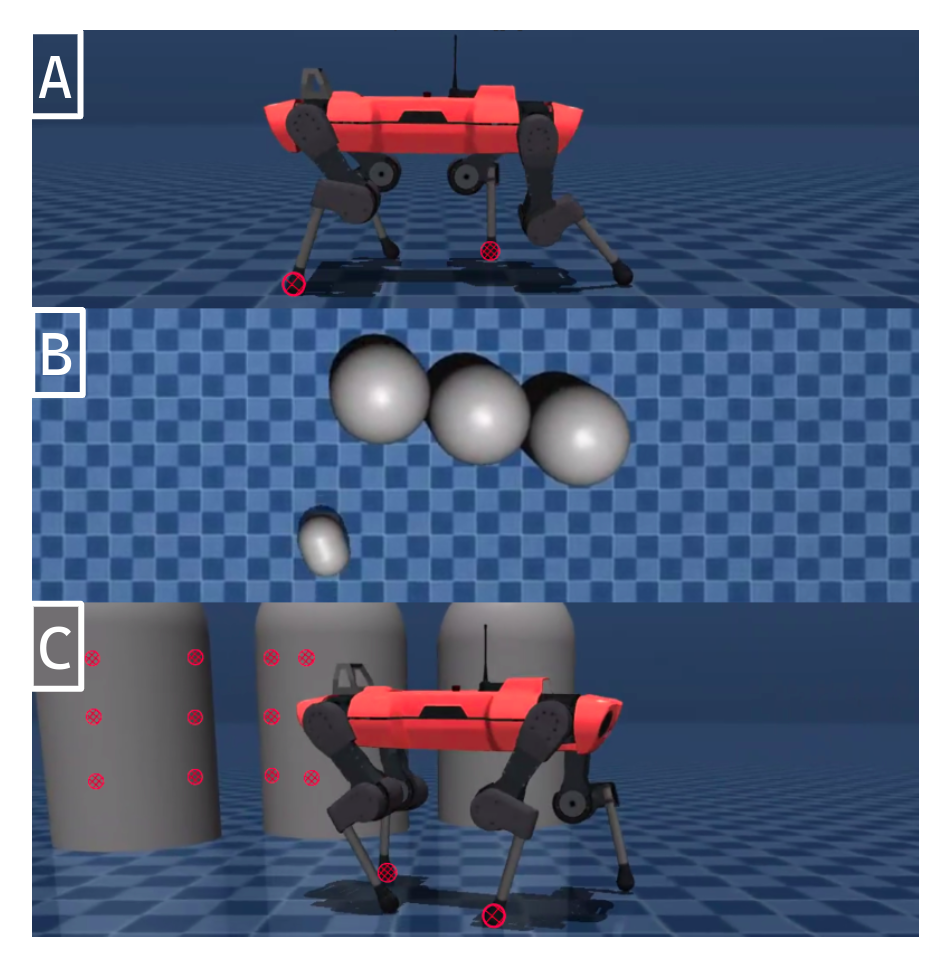}
    \caption{We apply differentiable simulation to learn A: walking, B: vision-based local navigation, C: walking around obstacles - tasks involving ground contact and depth rendering (depicted red). Analytical first-order policy gradients train A and C in minutes and B in seconds. \color{blue}{\href{https://m.youtube.com/watch?v=NcmkAH_nwvw}{Video}}, \color{blue}{\href{https://github.com/google-deepmind/mujoco/blob/main/mjx/training_apg.ipynb}{Locomotion code}}
}
    \label{fig:cover_pic}
\end{figure}

A new generation of powerful differentiable robot simulators, built on auto-differentiation frameworks, has fueled interest in using FoPGs for sample-efficient policy learning. FoPGs naturally suit tasks with smooth dynamics \cite{wiedemannTrainingEfficientControllers2023a, zhangBackNewtonsLaws2024, moraPODSPolicyOptimization2021} but optimizing in noisy landscapes, like those induced by surface contact, remains challenging. While previous approaches address this through algorithmic or simulation improvements, we exploit FoPGs' ability to converge accurately within local minima by adopting a residual policy learning (RPL) paradigm \cite{silverResidualPolicyLearning2019, johanninkResidualReinforcementLearning2019a}. In RPL, we guide the policy search with an anchor policy which we gradually augment with a learned residual. This is similar to hot-starting a trajectory optimizer by selecting an initial trajectory and hence a favorable local minimum for the motion search \cite{kellyIntroductionTrajectoryOptimization2017}. This strategy harnesses FoPGs' accuracy and sample efficiency while mitigating their inherent limitations in noisy loss landscapes.

We propose the following contributions:
\begin{itemize}
    \item We introduce residual policy learning with First-Order Policy Gradients (FoPGs), and compare their learning efficiency against Zeroth-Order Policy Gradients (ZoPGs).
    \item We provide novel insights on solving perceptive local navigation using FoPG's, enabling policy learning for a point-mass robot in seconds.
    \item We showcase  FoPG-RPL's versatility in jointly learning locomotion and perceptive navigation for a quadruped - a high-dimensional task with a complex loss landscape influenced by both contact dynamics and pixel inputs.
\end{itemize}

In this work, we directly auto-differentiate from the roll-out returns to the policy parameters, with gradients flowing through multiple steps of the full simulation physics in the process. This reduces engineering effort and provides more comprehensive gradient information compared to works that differentiate through heuristically simplified world models for policy training \cite{yangIPlannerImperativePath2023, amosDifferentiableMPCEndend2019}.

\section{Related Work}
\label{section:related}

\textbf{Deep Reinforcement Learning} Following the introduction of PPO \cite{schulmanProximalPolicyOptimization2017}, a seminal study \cite{heessEmergenceLocomotionBehaviours2017a} showed that RL using Zeroth-Order Policy Gradients (ZoPGs) can learn vision-based locomotion in obstacle-rich environments purely through environmental interaction, without requiring detailed reward shaping, training, or architecture design. Despite the long reported training times, similar approaches have been significantly accelerated with GPU-parallelized simulations, achieving tasks like blind locomotion within minutes \cite{rudinLearningWalkMinutes2022a}.

Most high-performance policies employ structured training processes to ensure robustness, style, and training convergence. Strategies include limiting the learning task to high-level foothold trajectories executed by an inverse-kinematics controller \cite{leeLearningQuadrupedalLocomotion2020}, constraining the search space to sinusoidal primitives \cite{bellegardaCPGRLLearningCentral2022c}, and guiding policy search with reference positions from optimal controllers \cite{jeneltenDTCDeepTracking2024}. Vision-based control often relies on additional heuristics: Teacher-Student learning trains the deployed policy (student) to imitate a teacher policy trained with more informative observations \cite{mikiLearningRobustPerceptive2022, chengExtremeParkourLegged2024}, while other approaches involve pre-processing the policy’s vision inputs \cite{heAgileSafeLearning2024}.

\textbf{Residual Policy Learning} Two concurrent works achieve robust object manipulation by simplifying policy search to learning a residual on top of a hand-crafted policy \cite{johanninkResidualReinforcementLearning2019a, silverResidualPolicyLearning2019}, utilizing low-variance off-policy RL gradients. Residual Policy Learning (RPL) can be seen as both a method for enhancing the robustness of a hand-designed controller through a learned residual and as a strategy to guide RL policy exploration using the hand-designed controller \cite{silverResidualPolicyLearning2019}. We extend this approach by employing analytical policy gradients to efficiently track detailed reward signals.

\textbf{Back-propagation through time (BPTT)} has demonstrated superior sample efficiency over Reinforcement Learning, particularly in contact-free tasks \cite{wiedemannTrainingEfficientControllers2023a, moraPODSPolicyOptimization2021, zhangBackNewtonsLaws2024}. Despite its advantages, BPTT faces challenges in contact-rich tasks like locomotion and manipulation, due to jagged loss landscapes \cite{antonovaRethinkingOptimizationDifferentiable2023} and poor gradient conditioning \cite{metzGradientsAreNot2022}. As shown in Table \ref{tab:checkmarks}, these issues are often mitigated by truncating trajectories. The Short-Horizon Actor-Critic algorithm \cite{xuAcceleratedPolicyLearning2022a} compensates for this by using a value function to approximate the rest of the infinite roll-out, enabling successful locomotion learning for high-DoF robots and robustness to contact stiffness \cite{schwarkeLearningQuadrupedalLocomotion2024a}. Alternatively, \cite{songLearningQuadrupedLocomotion2024} presents a stable BPTT training pipeline for quadruped locomotion, utilizing IsaacGym \cite{makoviychukIsaacGymHigh2021} to simulate forward dynamics while avoiding contact dynamics during the backward pass by separately simulating the legs and base.  BPTT's limitation of tending to converge to local minima serves as an advantage in motion mimicking, where the optimal state sequence is known and hence the learning process can be constantly reset to it \cite{renDiffMimicEfficientMotion2023}. Another frontier where BPTT has shown empirical promise is in vision-based control \cite{wiedemannTrainingEfficientControllers2023a, zhangBackNewtonsLaws2024}. We build on this line of work, providing further insights on the suitability of BPTT for pixel-based control.

\begin{table}[h]
    \centering
    \begin{tabular}{|>{\centering}m{8.5em}||>{\centering}m{3em}|>{\centering}m{4em}|>{\centering}m{4em}|>{\centering\arraybackslash}m{3em}|}
        \hline
        \textbf{Name} & Contact & Truncated trajectory & General Simulator & Vision \\ 
        \hhline{|==|=|=|=|}
        Brax APG \cite{freemanBraxDifferentiablePhysics2021} & \xmark & \checkmark & \checkmark & \xmark\\ 
        \hline
        PODS \cite{moraPODSPolicyOptimization2021} & \xmark & \checkmark & \xmark & \xmark \\ 
        \hline
        APG \cite{wiedemannTrainingEfficientControllers2023a} & \xmark & \checkmark & \xmark & \checkmark\\ 
        \hline
        Zhang et al. \cite{zhangBackNewtonsLaws2024} & \xmark & \xmark & \xmark & \checkmark\\
        \hline
        SHAC \cite{xuAcceleratedPolicyLearning2022a} & \checkmark & \checkmark & \checkmark & \xmark\\
        \hline
        Schwarke et al. \cite{schwarkeLearningQuadrupedalLocomotion2024a} & \checkmark & \checkmark & \xmark & \xmark\\
        \hline
        Song et al. \cite{songLearningQuadrupedLocomotion2024} & \checkmark & \checkmark & \xmark & \xmark\\
        \hline
        \textbf{FoPG RPL (ours)} & \checkmark & \checkmark & \checkmark & \checkmark\\
        \hline
    \end{tabular}
    \caption{Differentiable Simulation for Policy Learning}
    \label{tab:checkmarks}
\end{table}

As highlighted in Table \ref{tab:checkmarks}, our approach is uniquely applied to a learning problem that integrates both contact dynamics and perception. Unlike previous works that often require tailored modifications to the simulator for each robot, our use of a general articulated rigid body simulator (Mujoco MJX \cite{todorovMuJoCoPhysicsEngine2012}) ensures a straightforward and adaptable approach.

\section{Residual Policy Learning with FoPG's}

\begin{figure}
    \centering
    \includegraphics[width=1\linewidth]{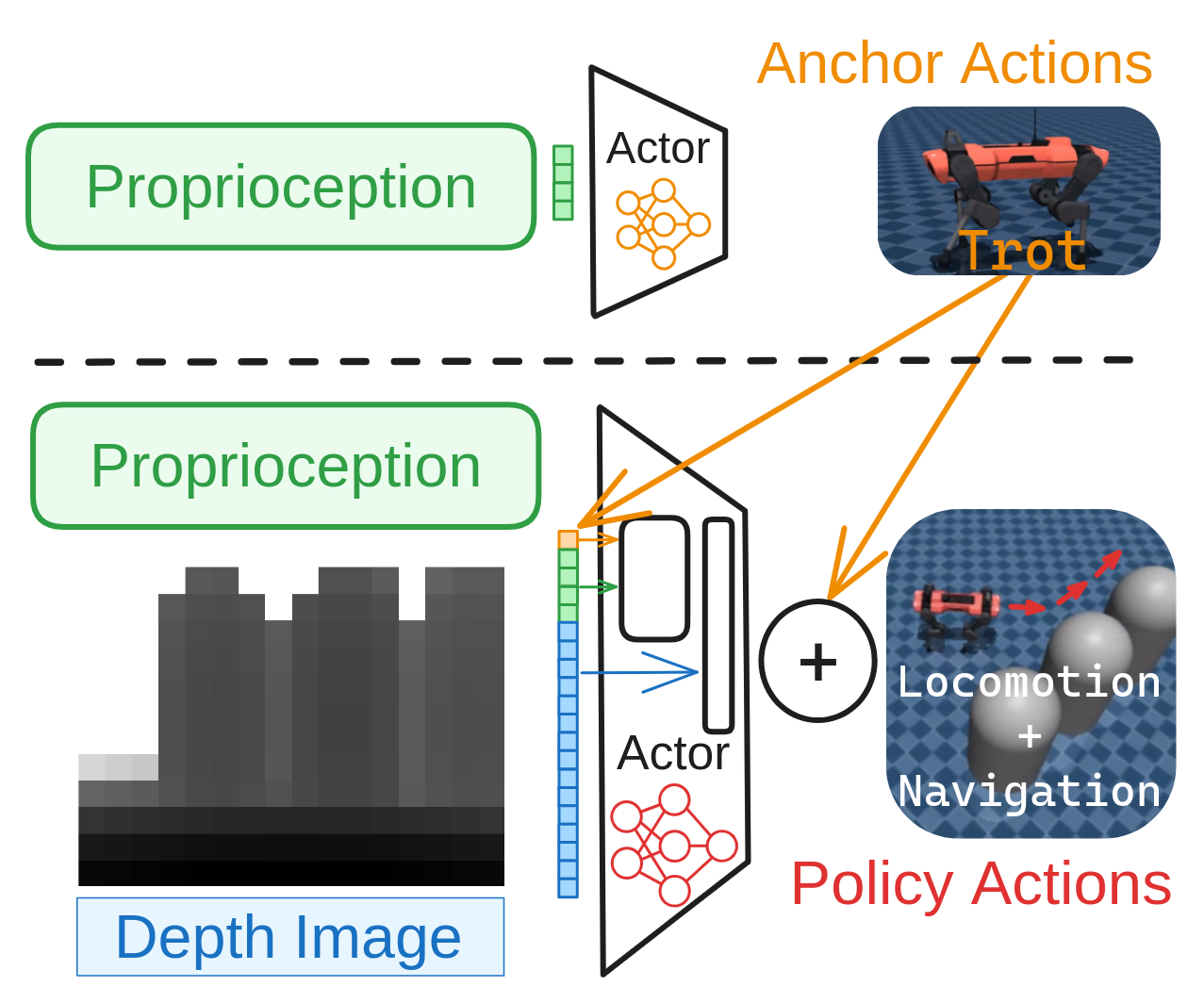}
    \caption{\textbf{Policy architecture for pixel to joint-angle perceptive navigation.} To guide learning the final policy, we use an \textcolor{orange}{anchor policy} with frozen weights (top) that is trained to nominally trot in place, with MLP layers of dimensions [256, 128]. Its output \emph{anchor actions} are both observed by the learned policy and added to the learned policy's outputs, similar to RPL \cite{silverResidualPolicyLearning2019, johanninkResidualReinforcementLearning2019a}. We pre-process the anchor actions alongside proprioceptive inputs with a dedicated MLP of dimensions [128, 64] before concatenating the flattened 16x12 depth camera input and applying a 128-neuron dense layer.}
    \label{fig:archi}
\end{figure}

In this section, we outline our methodology for achieving blind locomotion using FoPG's. This forms the state-based component of the perceptive locomotion problem depicted in Figure \ref{fig:archi}.

\subsection{Back-propagation through Time}
\label{subsection_BPTT}

BPTT methods differentiate through simulator computation graphs, enabling an analytical calculation of the policy gradient. Figure \ref{fig:pinocchio}A illustrates a case study in obstacle avoidance. This approach requires full differentiability, including within the simulator steps, observation calculations, and reward functions. As differentiating through each time step involves multiplying Jacobians \cite{metzGradientsAreNot2022}, BPTT gradients can become uninformative over long time sequences \cite{xuAcceleratedPolicyLearning2022a}. Consequently, prior works truncate the number of states used in gradient estimation \cite{songLearningQuadrupedLocomotion2024, renDiffMimicEfficientMotion2023, schwarkeLearningQuadrupedalLocomotion2024a}. In our work, we employ the Short-Horizon Actor-Critic (SHAC) algorithm, which extends truncated rollouts by approximating downstream rewards using a value function.

\subsection{Hot-starting and Anchor Learning}
\label{subsection:hot_starting}
One can \textit{hot-start} a policy search by learning a perturbation $\delta$ over a fixed baseline. Actions take the form $a_t \doteq b_t + \delta_{\theta,t}(x_t)$, where $\delta_{\theta,t}$ is a stochastic policy parameterized by $\theta$ with a small mean and variance uniformly across inputs $x_t$. This can be achieved in practice by initializing the neural network with small weight magnitudes. Hot-starting simplifies policy search by guiding exploration within the trajectory space near the baseline trajectory. The search progresses predictably when low-variance gradients and a small step size are used, allowing traversal down the local loss valley. This approach parallels Trajectory Optimization, where accurate loss gradients and non-convex loss landscapes make a good initial guess of the optimal trajectory $(x^0_0, a^0_0, ...,x^0_{T-1}, a^0_{T-1}, x^0_T)$ critical for convergence to a high-quality solution.

Setting $b_t$ to a cyclical hard-coded trot is sufficient for achieving a good learning outcome, as shown in Clip 2 of the supplementary video. However, due to limited PD gains, achieving faster trotting and enhanced gait stability requires closed-loop control. In \textit{Residual Policy Learning} (RPL), we replace the open-loop sequence with a closed-loop \textit{anchor policy} with frozen weights, enabling a faster trotting baseline. The goal of the policy search is to learn a \textit{residual policy} $\delta_{\theta, t}$ that augments the anchor $b_{\phi, t}$:

$$
a_t = b_{\phi, t}(x_t) + \delta_{\theta,t}(b_t, x_t).
$$

Like in Residual Learning for deep neural networks \cite{heDeepResidualLearning2016}, RPL preconditions a group of neural network layers to the identity transformation using a skip connection. If $a^*\sim\pi^*$ is the optimal action, the goal is to find some $\delta^*_\theta$ that outputs residuals $\delta^*_{\theta, t} = a^*_t - b_t$. In the case that $b_t$ are good actions, the augmentation policy $\delta$ ideally would not change them. Using the skip-connection from the baseline policy to the output, we precondition the learning of an identity augmentation such that it involves learning $\delta^* = 0$ rather than $a_t = b_t$, since learning the latter using a sequence of nested non-linear functions is intuitively more involved for an optimiser than driving outputs to zero.

RPL can be viewed both as augmenting the anchor policy and as guiding exploration. We term RPL using FoPG's as \textit{Anchor Learning} due to the role of the frozen policy in setting the local minimum.

\subsection{Training Details}
\label{subsection:loco_training_details}

We outline the training setup for blind locomotion on the Anybotics Anymal C \cite{hutterANYmalHighlyMobile2016} using the Mujoco \cite{todorovMuJoCoPhysicsEngine2012} MJX simulator. Our approach employs a standard state-based observation set for quadrupedal locomotion \cite{songLearningQuadrupedLocomotion2024}, but employs running-average observation normalisation. In addition to conventional reward terms such as velocity tracking and effort penalization \cite{schwarkeLearningQuadrupedalLocomotion2024a, rudinLearningWalkMinutes2022a}, we generate reference foot position trajectories using the Raibert Heuristic \cite{songLearningQuadrupedLocomotion2024}. While explicitly specifying foot target positions increases reward complexity and restricts the gait optimization search space, this method can reduce engineering effort compared to the typical RL approach, where designing a gait often involves iterating on high-level reward functions.

The residual policy uses a multi-layer perceptron with hidden layers of sizes [128, 64, 128]. The underlying anchor policy, which trots in place with each foot contacting the ground twice per second, is learned using APG to imitate hand-designed kinematics (sinusoidal trotting), similar to \cite{renDiffMimicEfficientMotion2023}. To encourage the residual policy to stay close to the anchor policy, we initialize the final layer's weights with magnitudes near zero \cite{silverResidualPolicyLearning2019}. We reduce the variance of the sampled actions by a factor of ten and find that layer normalisation \cite{baLayerNormalization2016} greatly improves training stability. We use hyperparameters from \cite{xuAcceleratedPolicyLearning2022a} for SHAC but decrease the learning rate to 1.5e-4.

% Arxiv doesn't like this part.
% \subsection{Code}
% An implementation of Section \ref{subsection:loco_training_details} is available in the Mujoco thub repository (:

% \vspace{0.3cm}
% \texttt{https://github.com/google-deepmind/mujoco}
% \vspace{1cm}

\section{Learning Local Navigation}

We now outline our methodology for using FoPG's to learn perceptive local navigation, where a robot tracks a body-frame forward velocity while using depth-camera inputs to avoid static obstacles within its field of view. This forms the vision-based component of the perceptive locomotion problem from Figure \ref{fig:archi}. We firstly isolate the vision-based problem by studying a trivial planar point-mass robot with two independently actuated degrees of freedom: one for heading and one for forward/backward movement. This robot moves like a differential drive, serving as an approximation of quadruped movement. A low-level velocity controller is employed to abstract away the task of learning the dynamics.

\subsection{Differentiating Through Pixels}
\label{section:diff_thru_pixels}

The analytical gradient of the returns with respect to the policy parameters includes the term 

\begin{equation}
\frac{da_t}{d\theta} = \frac{\partial a_t}{\partial \theta} + \left( \frac{\partial a_t}{\partial o_t} \frac{\partial o_t}{\partial x_t} + \frac{\partial a_t}{\partial v_t} \frac{\partial v_t}{\partial x_t} \right) \frac{\partial x_t}{\partial \theta},
\label{action_deriv}
\end{equation}

where $a_t = \pi_\theta(o_t, v_t)$ denotes the policy depending on its parameters $\theta$, state observations $o_t$, and depth camera inputs $v_t$. Notably, the term $\frac{\partial v_t}{\partial x_t}$ is the Jacobian of the rendering function. While differentiating through the renderer might seem necessary at first, previous works have successfully learned policies using BPTT without this term \cite{wiedemannTrainingEfficientControllers2023a, zhangBackNewtonsLaws2024}. Beyond implementation convenience, avoiding this term spares the computational cost of the $\mathcal{O}(nwh)$ Jacobian, where $n$ denotes the state dimensionality, and $w$ and $h$ are the image dimensions. We hypothesize that this term does not aid training because it intuitively guides actions to better position downstream vision measurements, such as gathering more information in the field of view. However, for our local-navigation task, this information is irrelevant as the critical data is nearly always within the field of view. Additionally, similar to ground contacts, poorly conditioned depth gradients near object edges are expected to degrade any learning signal.

\subsection{The Manoeuvre Window}
\label{subsection:ptrick}
Black-box ZoPGs can suffer from delayed feedback in local navigation \cite{rudinAdvancedSkillsLearning2022a}, lacking a mechanistic way to connect collisions with preceding actions. By using local physics to link preceding states and actions with sparse events such as obstacle collisions, BPTT alleviates this issue of sparse rewards. When the robot collides with an obstacle in the $h$-step forward pass, the BPTT gradient associates the collision with actions taken up to $h$ steps prior. This is referred to as the Manoeuvre Window \cite{zhangBackNewtonsLaws2024}. A straightforward approach to extend the Manoeuvre Window is to use longer trajectories, but this can result in poor gradient conditioning \cite{metzGradientsAreNot2022}.

\begin{figure}
    \centering
    \includegraphics[width=1\linewidth]{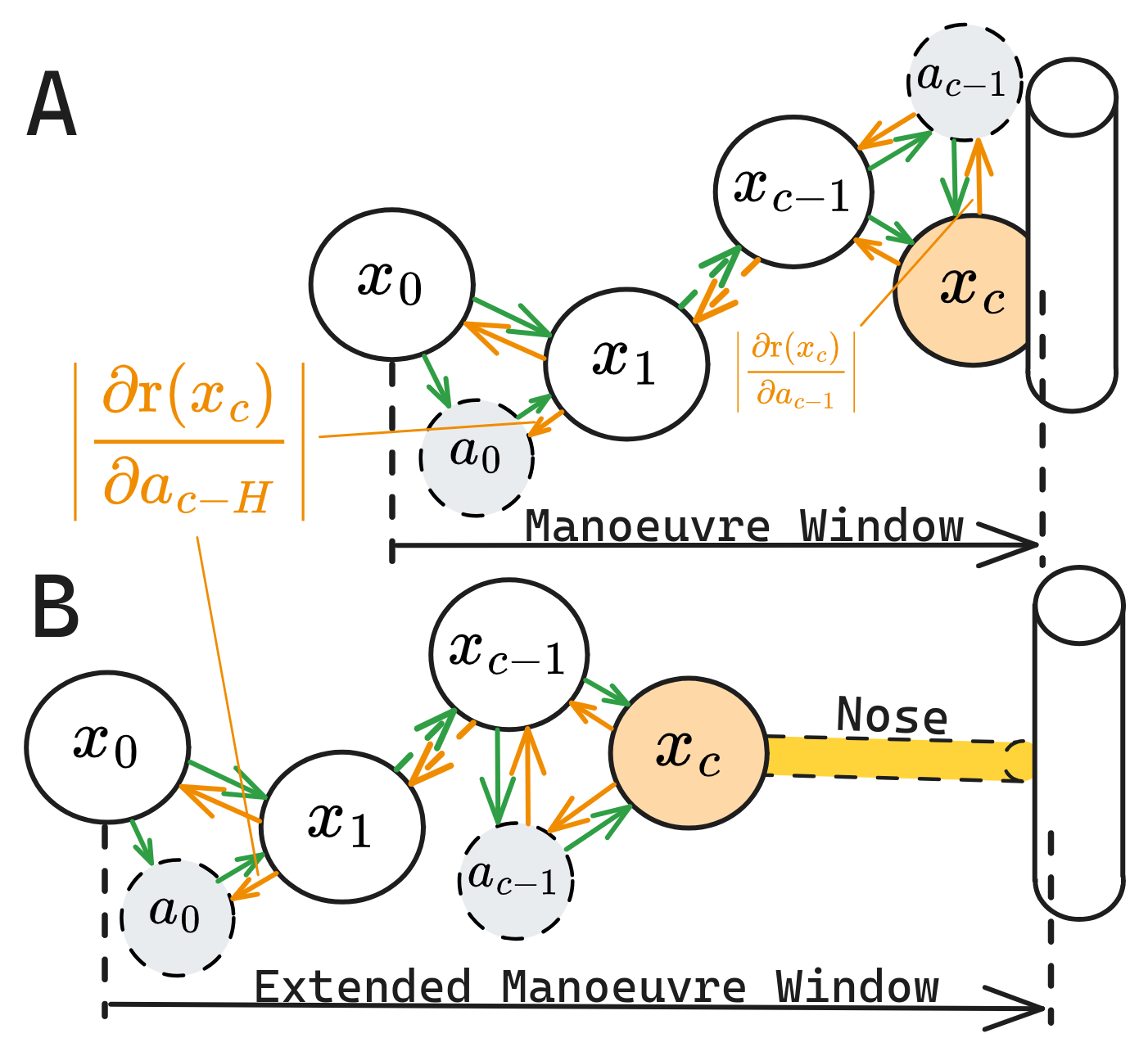}
    \caption{\textbf{The Pinnochio Trick.} Green denotes forward simulation, while yellow indicates gradient flow from the collision event. A: BPTT policy gradient calculation upon collision: the reward signal propagates back from the current state (highlighted in yellow) to actions up to $H$ time steps prior, allowing maneuvers to be learned within this length-$H$ window. Gradients increase in magnitude due to repeated multiplication with the simulation Jacobian. B: To extend the maneuver window without enlarging Jacobian sizes, we penalize collisions with a virtual \textit{nose} rigidly fixed to the robot's center of mass.}

    \label{fig:pinocchio}
\end{figure}

The local navigation task benefits from highly predictable future positions. We introduce the \textit{Pinocchio Trick}, which penalizes contact with a virtual \textit{nose} rigidly fixed to the robot base, rather than penalizing contact with the robot body itself. Figure \ref{fig:pinocchio} shows how by penalizing this projected location, the Manoeuvre Window is effectively extended without increasing the back-propagation burden through additional time steps.

Obstacle inflation is a common technique in path planning \cite{hanEnhancedAdaptive3D2023} which involves artificially enlarging obstacles in training to encourage conservative pathing during deployment. As will be discussed in the results, the Pinocchio Trick's extended manoeuvre window substantially improves learning dynamics. Obstacle inflation does not offer this benefit of an extended manouevre window, as the robot is simply tasked with avoiding a larger obstacle.

\subsection{Training Details}
\label{section:localnavdets}
We base our rewards on \cite{zhangBackNewtonsLaws2024} and similarly find benefit in smoothing the velocity observation. We replace the moving average with an exponential moving average, defined as $y_t = \alpha v_t + (1-\alpha) y_{t-1}$ with $y_0 = 0$, to improve training stability. The robot observes a 16x12 wide depth-camera input along with its velocity. We use custom point mass dynamics instead of the general-purpose Mujoco dynamics to improve simulation speed. The policy architecture consists of a single dense layer that connects the concatenated and flattened depth input and state observations to the output logits.

\section{Results}

In this section, we demonstrate the superior training convergence of our approach across three tasks: blind locomotion, local navigation and quadrupedal local navigation. We further explore two key topics from the FoPG literature: whether the learned policies are robust and how differentiable rendering impacts training dynamics. Statistics are calculated over five training runs and wall-clock times are reported based on the Nvidia 3060 TI GPU. The supplementary video is available at \texttt{https://youtu.be/NcmkAH\_nwvw}

\subsection{Residual Policy Learning for Blind Locomotion}
\label{subsection:results_blind_loco}
\begin{figure}
    \centering
    \includegraphics[width=1\linewidth]{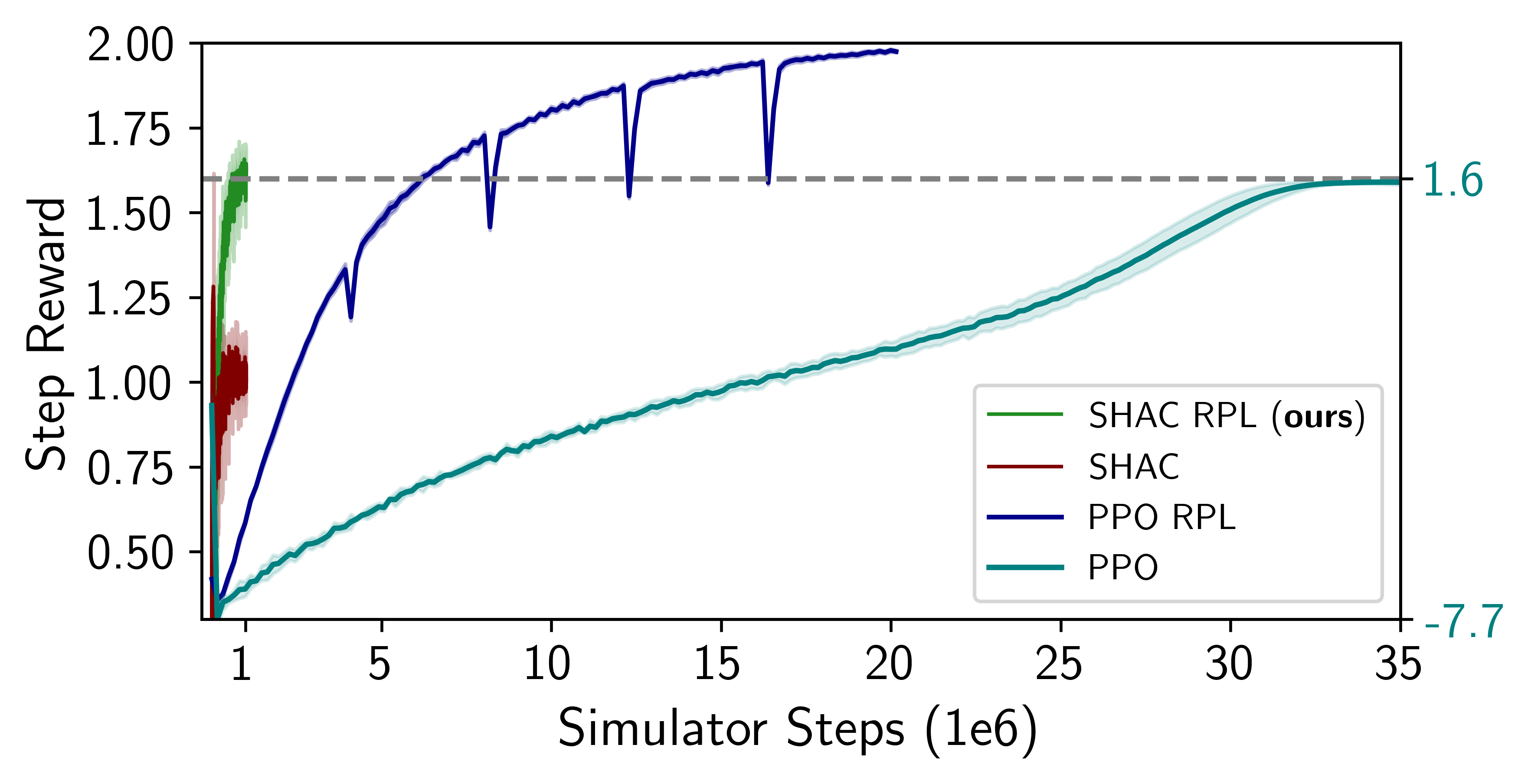}
    \caption{\textbf{Learning blind locomotion.} Residual Policy Learning increases asymptotic rewards for SHAC and PPO by 60\% and 25\% respectively while also improving the latter's sample efficiency by 6x. We plot PPO on a re-scaled y-axis for visualisation.}
    \label{fig:benchmarks}
\end{figure}

In Figure \ref{fig:benchmarks}, we study blind locomotion at 1 m/s (Section \ref{subsection:loco_training_details}) to compare the role of RPL in FoPG and ZoPG-based learning. In this experiment, we find both that FoPG-based learning benefits from the RPL framework and that the RPL framework benefits from using FoPG's in place of ZoPG's.

As hypothesized by the hot-starting interpretation from Section \ref{subsection:hot_starting}, we find that our approach of applying SHAC over an anchor policy (SHAC RPL) significantly improves the outcome policy. However, it does not appear to improve sample efficiency. Two possible reasons are the inherent sample-efficiency of FoPG's and the task difficulty in gradually learning to counteract the compensation strategies of the underlying closed-loop anchor policy.

We adapt Brax \cite{freemanBraxDifferentiablePhysics2021} implementations of SAC and PPO to the RPL framework by initialising the weights of the last neural network layer to small values \cite{silverResidualPolicyLearning2019} and decreasing the action sampling variance by a fixed factor of 10. Table \ref{tab:hyperparameters} shows the parameters for PPO RPL. While prior RPL works use off-policy RL for reduced sample variance, we find that SAC, a SOTA off-policy method, fails to learn in our RPL formulation. PPO adapts well to the RPL task, reflecting the general shift in quadruped locomotion research from sample-efficient off-policy methods to on-policy methods used in conjunction with accelerated data collection \cite{makoviychukIsaacGymHigh2021,rudinLearningWalkMinutes2022a}. Like in related works \cite{xuAcceleratedPolicyLearning2022a, songLearningQuadrupedLocomotion2024, schwarkeLearningQuadrupedalLocomotion2024a}, we find that FoPG's contribution to the RPL formulation lies mainly in improved sample efficiency. SHAC RPL is 35x and 6x more sample efficient than vanilla PPO and PPO RPL respectively. However, the impressive training dynamics of PPO RPL hint both that PPO's large batch-size of 163,840 samples is sufficient to accurately estimate policy gradients for this task and that BPTT methods may be inherently limited in policy exploration.

Comparing PPO and PPO RPL suggests different roles for RPL within ZoPG and FoPG-based training. Whereas the anchor policy benefits ZoPG with improved sample efficiency and asymptotic reward, the inherent sample efficiency of FoPG-based training renders the primary benefit of the anchor policy as setting a favourable local loss landscape.

\subsection{BPTT for Local Navigation}
\label{subsection:results_vision}
\begin{figure}
    \centering
    \includegraphics[width=1\linewidth]{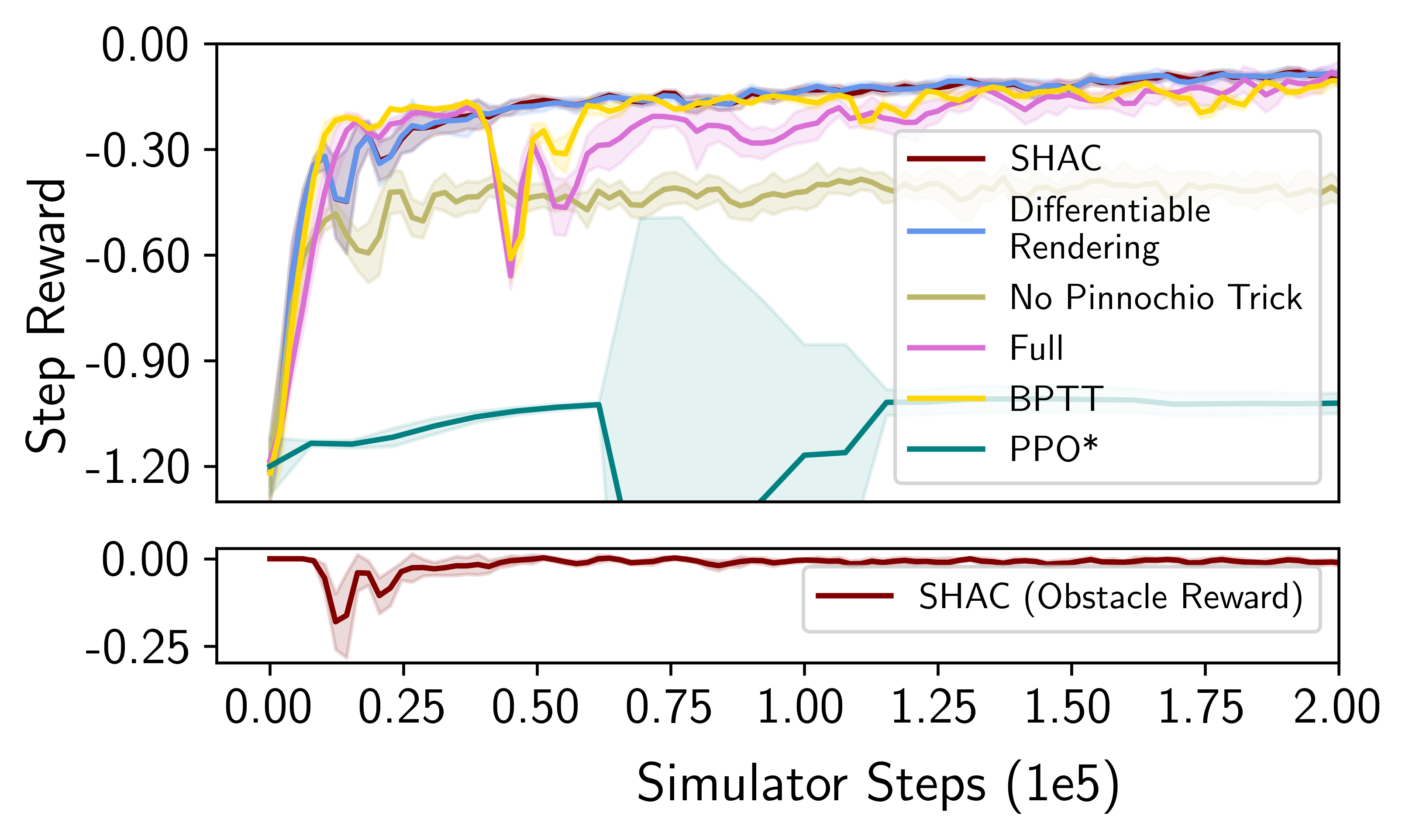}
    \caption{\textbf{Obstacle avoidance on a point mass robot.} Training the local navigation problem from Section \ref{section:localnavdets} using \emph{SHAC}, we test training convergence when adding \emph{differentiable rendering}, removing the \emph{Pinnochio Trick} or learning on a \emph{full}-scale vision setup with roughly 1000x more policy parameters. We also train using vanilla truncated \emph{BPTT} \cite{songLearningQuadrupedLocomotion2024}, converging similarly to SHAC within 2e5 samples; under two seconds wall-clock time. PPO struggles to converge compared to the FoPG methods. \emph{Note*}: We run \emph{PPO} for 1.5e8 steps and downscale the x-axis by 820 for visualization.}
    \label{fig:exp_3}
\end{figure}

In Figure \ref{fig:exp_3}, we isolate the vision problem (Section \ref{section:localnavdets}) to test our hypotheses of FoPG's usage in the local navigation task and to test whether PPO's success in the state-based setting transfers. We do not apply RPL due to the action space's simplicity. SHAC solves the pixel-based problem in seconds, learning to avoid obstacles within 50,000 simulator steps. We find that a vanilla BPTT method \cite{songLearningQuadrupedLocomotion2024} trains similarly, confirming that the training performance is due more to the inherent suitability of FoPG's for local navigation proposed in Section \ref{subsection:ptrick} rather than SHAC's algorithmic innovations. The importance of a large manoeuvre window is supported by the necessity of the Pinocchio trick for stable training. To show that our reported sample efficiency does not rely on the down-sampled image and minimal architecture, we train to a similar effect on a classic CNN-based Atari image preprocessing architecture applied to depth maps at 84x84 resolution \cite{schulmanProximalPolicyOptimization2017, mnihPlayingAtariDeep2013}. 

We apply PPO with hyperparameters adapted from \cite{gelesDemonstratingAgileFlight2024} and corroborate previous findings \cite{gelesDemonstratingAgileFlight2024, rudinAdvancedSkillsLearning2022a} that the standard method struggles to converge, even with prolonged training. This implies that simply averaging more roll-outs at the scale of a state-based problem is not sufficient to scale PPO to learning pixel-based control.

\subsection{Residual Policy Learning for Quadruped Local Navigation}

\begin{figure}
    \centering
    \includegraphics[width=1\linewidth]{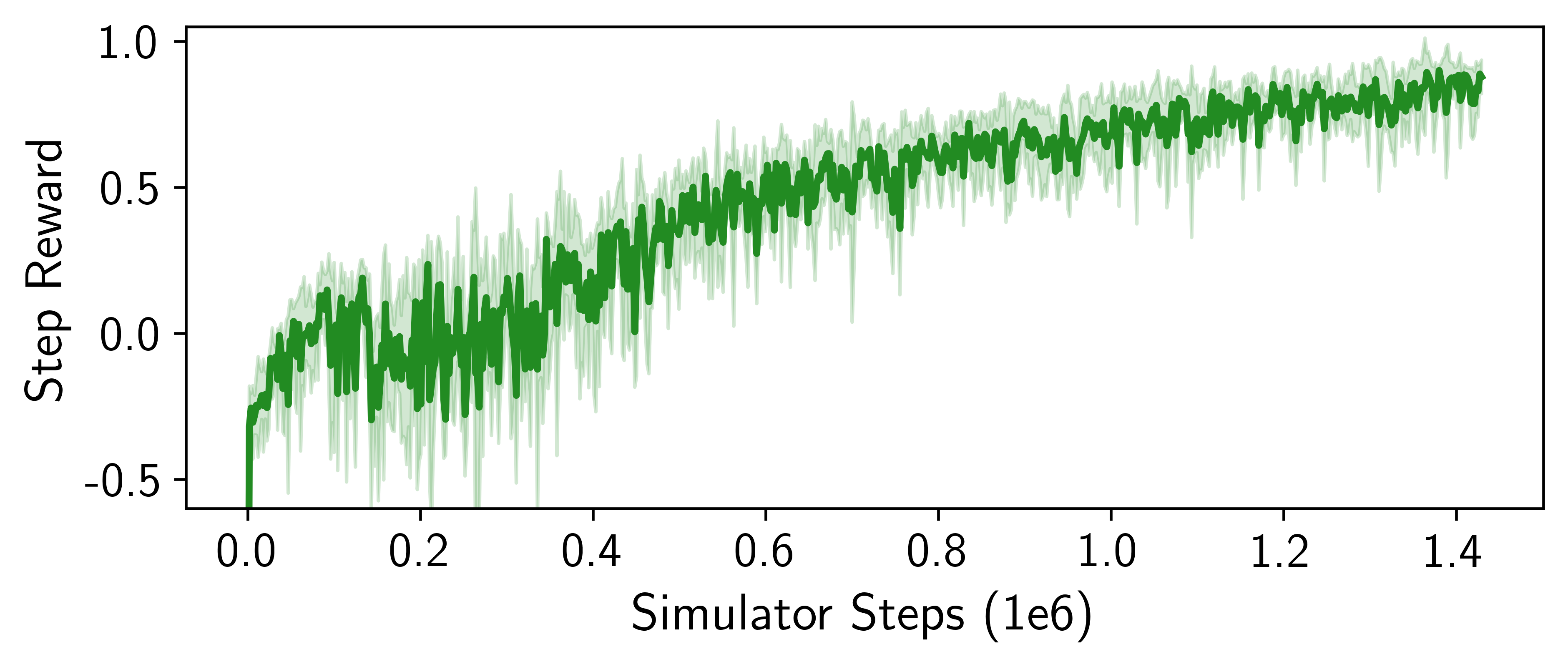}
    \caption{\textbf{Learning perceptive locomotion.} SHAC RPL trains Pixel-to-joint-angle local navigation on a quadruped with similar sample efficiency to blind locomotion (Figure \ref{fig:benchmarks}). 1.5e6 samples corresponds to 15 minutes of wall-clock time.}
    \label{fig:exp_4}
\end{figure}

Figure \ref{fig:exp_4} shows that SHAC RPL converges on the combined image and state-space control task from Figure \ref{fig:archi} with a similar number of samples as required for the state-only task. This expands upon previous findings of SHAC's suitability in high-dimensional settings \cite{xuAcceleratedPolicyLearning2022a}. The policy successfully learns to navigate around obstacles without explicit training for angular-velocity tracking. It is notable that the simple vision component in the policy architecture is sufficient to drive quadrupedal obstacle avoidance. Omitting a dedicated architecture for pre-processing the depth image may be possible due to its sufficiently informative representation. Since the pre-processing network for state-based observations cannot access obstacle information, the policy's success implies that the single dense layer applied after concatenating the depth image to the state embeddings is sufficient to transform straight locomotion into turning. As foreshadowed by section \ref{subsection:results_vision}, PPO and PPO RPL fail to train the task within 1e8 samples.

\subsection{Robustness of FoPG Policies}

\begin{figure}
    \centering
    \includegraphics[width=1\linewidth]{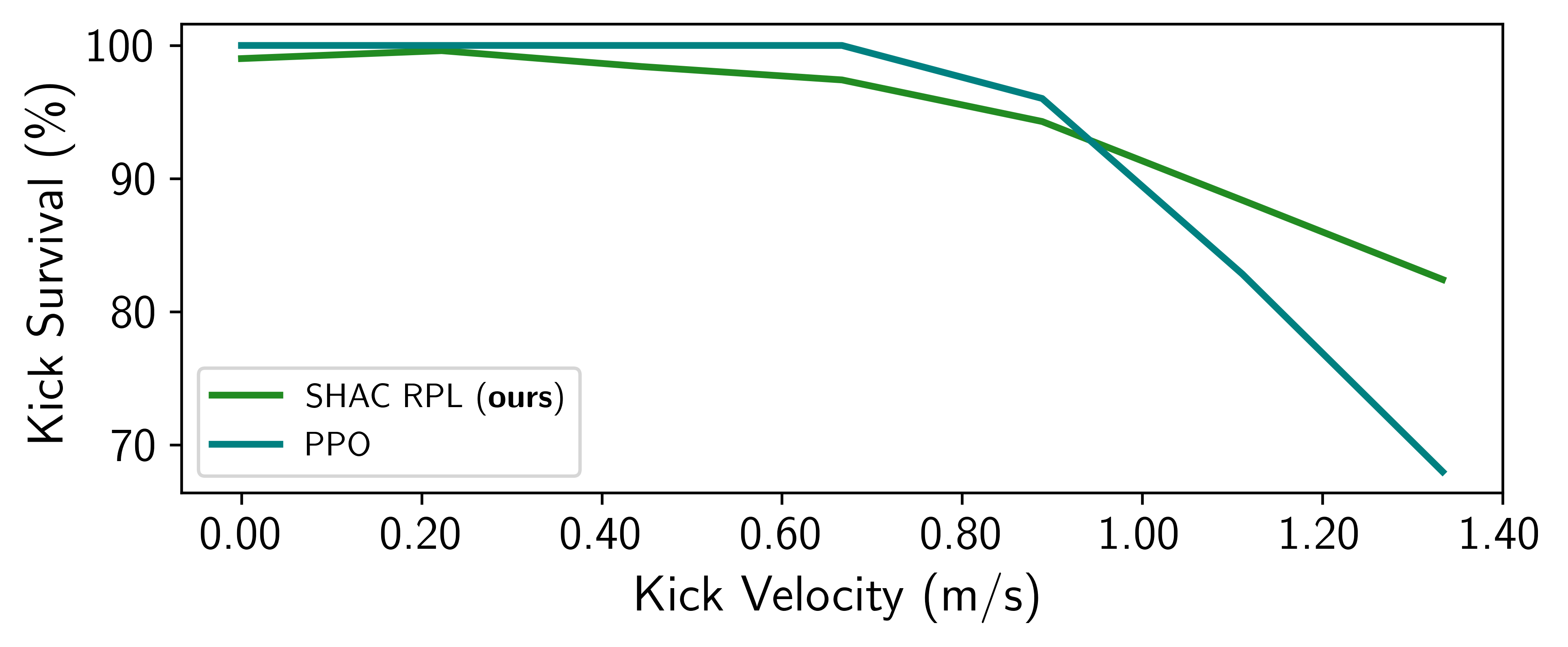}
    \caption{\textbf{Locomotion robustness.} SHAC RPL produces policies with robustness comparable to PPO when exposed to untrained disturbances. We simulate kicks by setting the robot base to a random velocity every two seconds. Percentages are calculated over 512 simulated kicks.}

    \label{fig:survival_rates}
\end{figure}
Given that FoPG-based methods require full differentiability and hence cannot benefit from certain task-level rewards such as staying upright, policy robustness naturally comes into question. In Figure \ref{fig:survival_rates}, we test the SHAC RPL policy from Section \ref{subsection:results_blind_loco} and find that it demonstrates comparable robustness against large, untrained disturbances with one trained with PPO, following the approach in \cite{rudinLearningWalkMinutes2022a}. The intuitive benefits of the non-differentiable survival reward manifest in the PPO policy's perfect survival for small perturbations. This aligns with previous findings on the importance of RL's task-level rewards \cite{songLearningQuadrupedLocomotion2024, songReachingLimitAutonomous2023}.

The SHAC point-mass local navigation policy achieves over 99\% obstacle avoidance within 100,000 simulator steps. The policy remains robust to Gaussian noise in depth readings, maintaining over 98\% obstacle avoidance and near-perfect velocity tracking even when unbiased Gaussian noise is scaled up to one meter. 

\subsection{Differentiable Rendering}
In Figure \ref{fig:exp_3}, we find that differentiating through the depth rays does not significantly affect SHAC's training outcomes, supporting our hypothesis from Section \ref{section:diff_thru_pixels}.

\section{Conclusion}

In this work, we jointly train locomotion and perceptive navigation on a quadruped by using FoPG's for residual policy learning. We isolate the quadrupedal locomotion and perceptive navigation components for further study. In the former, we show that the primary role of RPL for FoPG learning is to improve asymptotic rewards, whereas it can be used in ZoPG contexts for improved sample efficiency. Studying local navigation for a point-mass robot, we improve training by clipping the renderer gradients and penalising \emph{future} positions, demonstrating that the perceptive task can be solved in seconds.

There are several avenues for future work. The superior asymptotic reward of PPO RPL for blind locomotion indicates that our method could be enhanced by replacing SHAC with more a sophisticated learning algorithm such as AHAC \cite{georgievAdaptiveHorizonActorCritic2024a} or by utilizing another differentiable simulator with potentially higher gradient quality. Additionally, the local navigation task on a quadruped converges to a local minimum, resulting in lower robustness compared to the individual blind locomotion and local navigation tasks. Modularizing the training process could likely improve the final performance.

\section*{Acknowledgment}

Damien Berriaud revised the first draft and his insights lead to using momentum for velocity buffering and many ideas yet to be implemented. A conversation with Clemens Schwarke led to the guess that the depth camera inputs in local navigation do not require pre-processing. I thank Mohamed Hamoud for his incisive thinking throughout our many discussions.

\section{Hyper-parameters}

We present the parameters for PPO RPL, which effectively uses ZoPG's in the RPL framework for blind locomotion.

\begin{table}[h]
    \centering
    \renewcommand{\arraystretch}{1.1}
    \begin{tabular}{l c}
        \hline
        \textbf{Hyperparameter names} & \textbf{Values} \\
        \hhline{==}
        unroll length & 20 \\
        \hline
        number of minibatches & 32 \\
        \hline
        batch size & 256 \\
        \hline
        updates per batch & 4 \\
        \hline
        number of parallel environments & 4096 \\
        \hline
        learning rate & LinearDecay(3e-4, 1e-4) \\
        \hline
        entropy cost & 0.01 \\
        \hline
        % GAE $\lambda$ & 0.95 \\
        % \hline
        discounting & 0.97 \\
        \hline
    \end{tabular}
    \caption{Hyperparameters for PPO RPL}
    \label{tab:hyperparameters}
\end{table}
\balance

\bibliographystyle{IEEEtranBST/IEEEtran}
\bibliography{bibliography}

\begin{thebibliography}{10}
\providecommand{\url}[1]{#1}
\csname url@rmstyle\endcsname
\providecommand{\newblock}{\relax}
\providecommand{\bibinfo}[2]{#2}
\providecommand\BIBentrySTDinterwordspacing{\spaceskip=0pt\relax}
\providecommand\BIBentryALTinterwordstretchfactor{4}
\providecommand\BIBentryALTinterwordspacing{\spaceskip=\fontdimen2\font plus
\BIBentryALTinterwordstretchfactor\fontdimen3\font minus \fontdimen4\font\relax}
\providecommand\BIBforeignlanguage[2]{{%
\expandafter\ifx\csname l@#1\endcsname\relax
\typeout{** WARNING: IEEEtran.bst: No hyphenation pattern has been}%
\typeout{** loaded for the language `#1'. Using the pattern for}%
\typeout{** the default language instead.}%
\else
\language=\csname l@#1\endcsname
\fi
#2}}

\bibitem{mikiLearningRobustPerceptive2022}
\BIBentryALTinterwordspacing
T.~Miki, J.~Lee, J.~Hwangbo, L.~Wellhausen, V.~Koltun, and M.~Hutter, ``Learning robust perceptive locomotion for quadrupedal robots in the wild,'' vol.~7, no.~62, p. eabk2822. [Online]. Available: \url{http://arxiv.org/abs/2201.08117}
\BIBentrySTDinterwordspacing

\bibitem{chengExtremeParkourLegged2024}
\BIBentryALTinterwordspacing
X.~Cheng, K.~Shi, A.~Agarwal, and D.~Pathak, ``Extreme {{Parkour}} with {{Legged Robots}},'' in \emph{2024 {{IEEE International Conference}} on {{Robotics}} and {{Automation}} ({{ICRA}})}, pp. 11\,443--11\,450. [Online]. Available: \url{https://ieeexplore.ieee.org/abstract/document/10610200}
\BIBentrySTDinterwordspacing

\bibitem{suhDifferentiableSimulatorsGive2022a}
\BIBentryALTinterwordspacing
H.~J. Suh, M.~Simchowitz, K.~Zhang, and R.~Tedrake, ``Do {{Differentiable Simulators Give Better Policy Gradients}}?'' in \emph{Proceedings of the 39th {{International Conference}} on {{Machine Learning}}}.\hskip 1em plus 0.5em minus 0.4em\relax PMLR, pp. 20\,668--20\,696. [Online]. Available: \url{https://proceedings.mlr.press/v162/suh22b.html}
\BIBentrySTDinterwordspacing

\bibitem{mnihPlayingAtariDeep2013}
V.~Mnih, K.~Kavukcuoglu, D.~Silver, A.~Graves, I.~Antonoglou, D.~Wierstra, and M.~Riedmiller. Playing {{Atari}} with {{Deep Reinforcement Learning}}.

\bibitem{songReachingLimitAutonomous2023}
\BIBentryALTinterwordspacing
Y.~Song, A.~Romero, M.~Müller, V.~Koltun, and D.~Scaramuzza, ``Reaching the limit in autonomous racing: {{Optimal}} control versus reinforcement learning,'' vol.~8, no.~82, p. eadg1462. [Online]. Available: \url{https://www.science.org/doi/abs/10.1126/scirobotics.adg1462}
\BIBentrySTDinterwordspacing

\bibitem{rudinLearningWalkMinutes2022a}
\BIBentryALTinterwordspacing
N.~Rudin, D.~Hoeller, P.~Reist, and M.~Hutter, ``Learning to {{Walk}} in {{Minutes Using Massively Parallel Deep Reinforcement Learning}},'' in \emph{Proceedings of the 5th {{Conference}} on {{Robot Learning}}}.\hskip 1em plus 0.5em minus 0.4em\relax PMLR, pp. 91--100. [Online]. Available: \url{https://proceedings.mlr.press/v164/rudin22a.html}
\BIBentrySTDinterwordspacing

\bibitem{schulmanProximalPolicyOptimization2017}
\BIBentryALTinterwordspacing
J.~Schulman, F.~Wolski, P.~Dhariwal, A.~Radford, and O.~Klimov. Proximal {{Policy Optimization Algorithms}}. [Online]. Available: \url{http://arxiv.org/abs/1707.06347}
\BIBentrySTDinterwordspacing

\bibitem{schulmanHighDimensionalContinuousControl2018}
\BIBentryALTinterwordspacing
J.~Schulman, P.~Moritz, S.~Levine, M.~Jordan, and P.~Abbeel. High-{{Dimensional Continuous Control Using Generalized Advantage Estimation}}. [Online]. Available: \url{http://arxiv.org/abs/1506.02438}
\BIBentrySTDinterwordspacing

\bibitem{haarnojaSoftActorCriticPolicy2018}
\BIBentryALTinterwordspacing
T.~Haarnoja, A.~Zhou, P.~Abbeel, and S.~Levine, ``Soft {{Actor-Critic}}: {{Off-Policy Maximum Entropy Deep Reinforcement Learning}} with a {{Stochastic Actor}},'' in \emph{Proceedings of the 35th {{International Conference}} on {{Machine Learning}}}.\hskip 1em plus 0.5em minus 0.4em\relax PMLR, pp. 1861--1870. [Online]. Available: \url{https://proceedings.mlr.press/v80/haarnoja18b.html}
\BIBentrySTDinterwordspacing

\bibitem{heAgileSafeLearning2024}
\BIBentryALTinterwordspacing
T.~He, C.~Zhang, W.~Xiao, G.~He, C.~Liu, and G.~Shi, ``Agile {{But Safe}}: {{Learning Collision-Free High-Speed Legged Locomotion}},'' in \emph{Proceedings of {{Robotics}}: {{Science}} and {{Systems}}}.\hskip 1em plus 0.5em minus 0.4em\relax {Robotics: Science and Systems Foundation}. [Online]. Available: \url{http://arxiv.org/abs/2401.17583}
\BIBentrySTDinterwordspacing

\bibitem{wiedemannTrainingEfficientControllers2023a}
\BIBentryALTinterwordspacing
N.~Wiedemann, V.~Wüest, A.~Loquercio, M.~Müller, D.~Floreano, and D.~Scaramuzza, ``Training {{Efficient Controllers}} via {{Analytic Policy Gradient}},'' in \emph{2023 {{IEEE International Conference}} on {{Robotics}} and {{Automation}} ({{ICRA}})}, pp. 1349--1356. [Online]. Available: \url{https://ieeexplore.ieee.org/abstract/document/10160581}
\BIBentrySTDinterwordspacing

\bibitem{xuAcceleratedPolicyLearning2022a}
\BIBentryALTinterwordspacing
J.~Xu, V.~Makoviychuk, Y.~Narang, F.~Ramos, W.~Matusik, A.~Garg, and M.~Macklin, ``Accelerated {{Policy Learning}} with {{Parallel Differentiable Simulation}},'' in \emph{Proceedings of the 10th {{International Conference}} on {{Learning Representations}} ({{ICLR}} 2022)}. [Online]. Available: \url{https://arxiv.org/abs/2204.07137v1}
\BIBentrySTDinterwordspacing

\bibitem{freemanBraxDifferentiablePhysics2021}
\BIBentryALTinterwordspacing
C.~D. Freeman, E.~Frey, A.~Raichuk, S.~Girgin, I.~Mordatch, and O.~Bachem. Brax -- {{A Differentiable Physics Engine}} for {{Large Scale Rigid Body Simulation}}. [Online]. Available: \url{http://arxiv.org/abs/2106.13281}
\BIBentrySTDinterwordspacing

\bibitem{antonovaRethinkingOptimizationDifferentiable2023}
\BIBentryALTinterwordspacing
R.~Antonova, J.~Yang, K.~M. Jatavallabhula, and J.~Bohg, ``Rethinking {{Optimization}} with {{Differentiable Simulation}} from a {{Global Perspective}},'' in \emph{Proceedings of {{The}} 6th {{Conference}} on {{Robot Learning}}}.\hskip 1em plus 0.5em minus 0.4em\relax PMLR, pp. 276--286. [Online]. Available: \url{https://proceedings.mlr.press/v205/antonova23a.html}
\BIBentrySTDinterwordspacing

\bibitem{zhangBackNewtonsLaws2024}
\BIBentryALTinterwordspacing
Y.~Zhang, Y.~Hu, Y.~Song, D.~Zou, and W.~Lin. Back to {{Newton}}'s {{Laws}}: {{Learning Vision-based Agile Flight}} via {{Differentiable Physics}}. [Online]. Available: \url{http://arxiv.org/abs/2407.10648}
\BIBentrySTDinterwordspacing

\bibitem{moraPODSPolicyOptimization2021}
\BIBentryALTinterwordspacing
M.~A.~Z. Mora, M.~Peychev, S.~Ha, M.~Vechev, and S.~Coros, ``{{PODS}}: {{Policy Optimization}} via {{Differentiable Simulation}},'' in \emph{Proceedings of the 38th {{International Conference}} on {{Machine Learning}}}.\hskip 1em plus 0.5em minus 0.4em\relax PMLR, pp. 7805--7817. [Online]. Available: \url{https://proceedings.mlr.press/v139/mora21a.html}
\BIBentrySTDinterwordspacing

\bibitem{silverResidualPolicyLearning2019}
\BIBentryALTinterwordspacing
T.~Silver, K.~Allen, J.~Tenenbaum, and L.~Kaelbling. Residual {{Policy Learning}}. [Online]. Available: \url{http://arxiv.org/abs/1812.06298}
\BIBentrySTDinterwordspacing

\bibitem{johanninkResidualReinforcementLearning2019a}
\BIBentryALTinterwordspacing
T.~Johannink, S.~Bahl, A.~Nair, J.~Luo, A.~Kumar, M.~Loskyll, J.~A. Ojea, E.~Solowjow, and S.~Levine, ``Residual {{Reinforcement Learning}} for {{Robot Control}},'' in \emph{2019 {{International Conference}} on {{Robotics}} and {{Automation}} ({{ICRA}})}, pp. 6023--6029. [Online]. Available: \url{https://ieeexplore.ieee.org/abstract/document/8794127}
\BIBentrySTDinterwordspacing

\bibitem{kellyIntroductionTrajectoryOptimization2017}
\BIBentryALTinterwordspacing
M.~Kelly, ``An {{Introduction}} to {{Trajectory Optimization}}: {{How}} to {{Do Your Own Direct Collocation}},'' vol.~59, no.~4, pp. 849--904. [Online]. Available: \url{https://epubs.siam.org/doi/10.1137/16M1062569}
\BIBentrySTDinterwordspacing

\bibitem{yangIPlannerImperativePath2023}
\BIBentryALTinterwordspacing
F.~Yang, C.~Wang, C.~Cadena, and M.~Hutter, ``{{iPlanner}}: {{Imperative Path Planning}},'' in \emph{Robotics: {{Science}} and {{Systems XIX}}}.\hskip 1em plus 0.5em minus 0.4em\relax {Robotics: Science and Systems Foundation}. [Online]. Available: \url{http://www.roboticsproceedings.org/rss19/p064.pdf}
\BIBentrySTDinterwordspacing

\bibitem{amosDifferentiableMPCEndend2019}
\BIBentryALTinterwordspacing
B.~Amos, I.~D.~J. Rodriguez, J.~Sacks, B.~Boots, and J.~Z. Kolter. Differentiable {{MPC}} for {{End-to-end Planning}} and {{Control}}. [Online]. Available: \url{http://arxiv.org/abs/1810.13400}
\BIBentrySTDinterwordspacing

\bibitem{heessEmergenceLocomotionBehaviours2017a}
\BIBentryALTinterwordspacing
N.~Heess, D.~TB, S.~Sriram, J.~Lemmon, J.~Merel, G.~Wayne, Y.~Tassa, T.~Erez, Z.~Wang, S.~M.~A. Eslami, M.~Riedmiller, and D.~Silver. Emergence of {{Locomotion Behaviours}} in {{Rich Environments}}. [Online]. Available: \url{http://arxiv.org/abs/1707.02286}
\BIBentrySTDinterwordspacing

\bibitem{leeLearningQuadrupedalLocomotion2020}
\BIBentryALTinterwordspacing
J.~Lee, J.~Hwangbo, L.~Wellhausen, V.~Koltun, and M.~Hutter, ``Learning {{Quadrupedal Locomotion}} over {{Challenging Terrain}},'' vol.~5, no.~47, p. eabc5986. [Online]. Available: \url{http://arxiv.org/abs/2010.11251}
\BIBentrySTDinterwordspacing

\bibitem{bellegardaCPGRLLearningCentral2022c}
\BIBentryALTinterwordspacing
G.~Bellegarda and A.~Ijspeert, ``{{CPG-RL}}: {{Learning Central Pattern Generators}} for {{Quadruped Locomotion}},'' vol.~7, no.~4, pp. 12\,547--12\,554. [Online]. Available: \url{https://ieeexplore.ieee.org/abstract/document/9932888}
\BIBentrySTDinterwordspacing

\bibitem{jeneltenDTCDeepTracking2024}
\BIBentryALTinterwordspacing
F.~Jenelten, J.~He, F.~Farshidian, and M.~Hutter, ``{{DTC}}: {{Deep Tracking Control}},'' vol.~9, no.~86, p. eadh5401. [Online]. Available: \url{https://www.science.org/doi/10.1126/scirobotics.adh5401}
\BIBentrySTDinterwordspacing

\bibitem{metzGradientsAreNot2022}
\BIBentryALTinterwordspacing
L.~Metz, C.~D. Freeman, S.~S. Schoenholz, and T.~Kachman. Gradients are {{Not All You Need}}. [Online]. Available: \url{http://arxiv.org/abs/2111.05803}
\BIBentrySTDinterwordspacing

\bibitem{schwarkeLearningQuadrupedalLocomotion2024a}
\BIBentryALTinterwordspacing
C.~Schwarke, V.~Klemm, J.~Tordesillas, J.-P. Sleiman, and M.~Hutter. Learning {{Quadrupedal Locomotion}} via {{Differentiable Simulation}}. [Online]. Available: \url{http://arxiv.org/abs/2404.02887}
\BIBentrySTDinterwordspacing

\bibitem{songLearningQuadrupedLocomotion2024}
\BIBentryALTinterwordspacing
Y.~Song, S.~Kim, and D.~Scaramuzza. Learning {{Quadruped Locomotion Using Differentiable Simulation}}. [Online]. Available: \url{http://arxiv.org/abs/2403.14864}
\BIBentrySTDinterwordspacing

\bibitem{makoviychukIsaacGymHigh2021}
\BIBentryALTinterwordspacing
V.~Makoviychuk, L.~Wawrzyniak, Y.~Guo, M.~Lu, K.~Storey, M.~Macklin, D.~Hoeller, N.~Rudin, A.~Allshire, A.~Handa, and G.~State, ``Isaac {{Gym}}: {{High Performance GPU-Based Physics Simulation For Robot Learning}},'' in \emph{Proceedings of the {{Neural Information Processing Systems Track}} on {{Datasets}} and {{Benchmarks}}}.\hskip 1em plus 0.5em minus 0.4em\relax {Conference and Workshop on Neural Information Processing Systems}. [Online]. Available: \url{http://arxiv.org/abs/2108.10470}
\BIBentrySTDinterwordspacing

\bibitem{renDiffMimicEfficientMotion2023}
\BIBentryALTinterwordspacing
J.~Ren, C.~Yu, S.~Chen, X.~Ma, L.~Pan, and Z.~Liu, ``{{DiffMimic}}: {{Efficient Motion Mimicking}} with {{Differentiable Physics}},'' in \emph{The {{Eleventh International Conference}} on {{Learning Representations}}, {{ICLR}} 2023}.\hskip 1em plus 0.5em minus 0.4em\relax arXiv. [Online]. Available: \url{http://arxiv.org/abs/2304.03274}
\BIBentrySTDinterwordspacing

\bibitem{todorovMuJoCoPhysicsEngine2012}
\BIBentryALTinterwordspacing
E.~Todorov, T.~Erez, and Y.~Tassa, ``{{MuJoCo}}: {{A}} physics engine for model-based control,'' in \emph{2012 {{IEEE}}/{{RSJ International Conference}} on {{Intelligent Robots}} and {{Systems}}}.\hskip 1em plus 0.5em minus 0.4em\relax IEEE, pp. 5026--5033. [Online]. Available: \url{http://ieeexplore.ieee.org/document/6386109/}
\BIBentrySTDinterwordspacing

\bibitem{heDeepResidualLearning2016}
\BIBentryALTinterwordspacing
K.~He, X.~Zhang, S.~Ren, and J.~Sun, ``Deep {{Residual Learning}} for {{Image Recognition}},'' in \emph{Proceedings of the {{IEEE Conference}} on {{Computer Vision}} and {{Pattern Recognition}} ({{CVPR}}), 2016}, pp. 770--778. [Online]. Available: \url{https://openaccess.thecvf.com/content_cvpr_2016/html/He_Deep_Residual_Learning_CVPR_2016_paper.html}
\BIBentrySTDinterwordspacing

\bibitem{hutterANYmalHighlyMobile2016}
\BIBentryALTinterwordspacing
M.~Hutter, C.~Gehring, D.~Jud, A.~Lauber, C.~D. Bellicoso, V.~Tsounis, J.~Hwangbo, K.~Bodie, P.~Fankhauser, M.~Bloesch, R.~Diethelm, S.~Bachmann, A.~Melzer, and M.~Hoepflinger, ``{{ANYmal}} - a highly mobile and dynamic quadrupedal robot,'' in \emph{2016 {{IEEE}}/{{RSJ International Conference}} on {{Intelligent Robots}} and {{Systems}} ({{IROS}})}, pp. 38--44. [Online]. Available: \url{https://ieeexplore.ieee.org/document/7758092}
\BIBentrySTDinterwordspacing

\bibitem{baLayerNormalization2016}
\BIBentryALTinterwordspacing
J.~L. Ba, J.~R. Kiros, and G.~E. Hinton. Layer {{Normalization}}. arXiv.org. [Online]. Available: \url{https://arxiv.org/abs/1607.06450v1}
\BIBentrySTDinterwordspacing

\bibitem{rudinAdvancedSkillsLearning2022a}
\BIBentryALTinterwordspacing
N.~Rudin, D.~Hoeller, M.~Bjelonic, and M.~Hutter, ``Advanced {{Skills}} by {{Learning Locomotion}} and {{Local Navigation End-to-End}},'' in \emph{2022 {{IEEE}}/{{RSJ International Conference}} on {{Intelligent Robots}} and {{Systems}} ({{IROS}})}, pp. 2497--2503. [Online]. Available: \url{https://ieeexplore.ieee.org/abstract/document/9981198}
\BIBentrySTDinterwordspacing

\bibitem{hanEnhancedAdaptive3D2023}
\BIBentryALTinterwordspacing
L.~Han, L.~He, X.~Sun, Z.~Li, and Y.~Zhang, ``An enhanced adaptive {{3D}} path planning algorithm for mobile robots with obstacle buffering and improved {{Theta}}* using minimum snap trajectory smoothing,'' vol.~35, no.~10, p. 101844. [Online]. Available: \url{https://www.sciencedirect.com/science/article/pii/S1319157823003981}
\BIBentrySTDinterwordspacing

\bibitem{gelesDemonstratingAgileFlight2024}
\BIBentryALTinterwordspacing
I.~Geles, L.~Bauersfeld, A.~Romero, J.~Xing, and D.~Scaramuzza, ``Demonstrating {{Agile Flight}} from {{Pixels}} without {{State Estimation}},'' in \emph{Proceedings of {{Robotics}}: {{Science}} and {{Systems}}}.\hskip 1em plus 0.5em minus 0.4em\relax {Robotics: Science and Systems Foundation}. [Online]. Available: \url{https://arxiv.org/abs/2406.12505v1}
\BIBentrySTDinterwordspacing

\bibitem{georgievAdaptiveHorizonActorCritic2024a}
\BIBentryALTinterwordspacing
I.~Georgiev, K.~Srinivasan, J.~Xu, E.~Heiden, and A.~Garg, ``Adaptive {{Horizon Actor-Critic}} for {{Policy Learning}} in {{Contact-Rich Differentiable Simulation}},'' in \emph{Proceedings of the {{Forty-first International Conference}} on {{Machine Learning}}}.\hskip 1em plus 0.5em minus 0.4em\relax PMLR. [Online]. Available: \url{https://arxiv.org/abs/2405.17784v2}
\BIBentrySTDinterwordspacing

\end{thebibliography}

\end{document}